%% file: iclr2026_conference.tex
\documentclass{article}
\usepackage{iclr2026_conference,times}
\input{math_commands.tex}

\usepackage{hyperref}
\usepackage{url}
\usepackage{booktabs}
\usepackage{amsmath}
\usepackage{graphicx}
\usepackage{xcolor}

\title{Disparities in Negation Understanding \\ Across Languages in Vision-Language Models}
\author{
  Charikleia Moraitaki$^{*}$ \quad
  Sarah Pan$^{*}$ \quad
  Skyler Pulling$^{*}$ \\
  \bf Gwendolyn Flusche$^{*}$ \quad
  Kumail Alhamoud \quad
  Marzyeh Ghassemi \\
  \\
  \normalfont Massachusetts Institute of Technology \\
  \\
  \normalfont \small{$^{*}$Equal contribution}
}

\iclrfinalcopy
\begin{document}

\maketitle
\lhead{Published as a workshop paper at the AFAA Workshop, ICLR 2026}

\begin{abstract}
Vision-language models (VLMs) exhibit \emph{affirmation bias}: a systematic tendency to select positive captions (``X is present'') even when the correct description contains negation (``no X''). While prior work has documented this failure mode in English and proposed solutions, negation manifests differently across languages through varying morphology, word order, and cliticization patterns--raising the question of whether these solutions serve all linguistic communities equitably. We introduce the first human-verified multilingual negation benchmark, spanning seven typologically diverse languages: English, Mandarin Chinese, Arabic, Greek, Russian, Tagalog, and Spanish. Evaluating three VLMs--CLIP, SigLIP, and MultiCLIP--we find that standard CLIP performs at or below chance on non-Latin-script languages, while MultiCLIP achieves the highest and most uniform accuracy. We also evaluate SpaceVLM, a proposed negation correction, and find that it produces substantial improvements for several languages--particularly English, Greek, Spanish, and Tagalog--while showing varied effectiveness across typologically different languages. This variation reveals that linguistic properties like morphology, script, and negation structure interact with model improvements in fairness-relevant ways. As VLMs are deployed globally, multilingual benchmarks are essential for understanding not just whether solutions work, but for whom.
\end{abstract}

\section{Introduction}
Prompts to language models are overwhelmingly given in the affirmative: ``generate an image with a dog,'' ``give me a list that contains recipes.'' What happens when we negate? Recent work by ~\citet{alhamoud2025negation} demonstrates that state-of-the-art vision-language models (VLMs) frequently fail on negated queries, exhibiting \emph{affirmation bias}: the tendency to match a caption like ``there is no boat'' to an image containing a boat, simply because the word ``boat'' appears. This failure has serious downstream consequences—in radiology, the distinction between ``no pleural effusion'' and ``pleural effusion present'' determines whether a finding is flagged or dismissed.

However, all existing evaluations of negation understanding focus exclusively on English. This is a significant gap because negation is not always expressed the same way. For example, English uses simple adverbial particles (``no,'' ``not''), but Greek employs verbal negation with existential constructions (``$\delta\varepsilon\nu$ $\upsilon\pi\acute{\alpha}\rho\chi\varepsilon\iota$'' = ``there does not exist''), Arabic uses cliticized markers in a right-to-left script, and Chinese relies on isolating particles with distinct semantic functions. These structural differences may influence VLM behavior in ways that English-only benchmarks cannot reflect.
This is crucial for fairness, because LLMs are disproportionately built around English and a small handful of other high-resource languages, creating a ``digital language divide'' that systematically disadvantages most of the world’s languages \citep{bella2023bridging}. 
Prior work has documented systematic performance inequalities across languages in NLP more broadly \citep{blasi2022systematic}, and if VLMs similarly handle negation better for some linguistic communities than others, this constitutes an inequity that deserves attention—particularly as these systems enter safety-critical domains.

Our primary contribution is the first multilingual negation benchmark for VLMs, spanning seven typologically diverse languages. Using this benchmark, we find that even explicitly multilingual VLMs exhibit significant cross-lingual negation gaps, with performance varying by up to 27.5 percentage points across languages. We further apply SpaceVLM \citep{ranjbar2025spacevlm}—a recently proposed negation correction—across all languages and uncover that the method's effectiveness correlates with how a language expresses negation morphologically. This variation is itself informative: it reveals that linguistic structure shapes model behavior in ways that inform equitable deployment.

\section{Approach}

\paragraph{Benchmark construction.} We extend the English NegBench dataset \citep{alhamoud2025negation}, which contains 5,914 image-caption sets from COCO \citep{lin2014coco} with 4-way multiple choice, to seven languages chosen for typological diversity: English, Spanish, Greek, and Tagalog (negation via independent particles), Russian and Arabic (morphologically complex negation), and Mandarin Chinese (isolating, with distinct negation particles for different semantic functions). These languages also span Latin and non-Latin scripts, left-to-right and right-to-left writing, and multiple language families. Translations used Google Translate followed by human verification from native speakers ($30$ samples/language), who checked that negation markers were correctly translated, that negated sentences remained semantically faithful to the English source, and that phrasing was natural in the target language.

\paragraph{Models and evaluation.} We evaluate three contrastive VLMs representing different training regimes: CLIP \citep{radford2021clip} (primarily English-trained), SigLIP \citep{zhai2023siglip} (multilingual), and MultiCLIP \citep{carlsson2022mclip} (multilingual). We also evaluate after applying SpaceVLM \citep{ranjbar2025spacevlm}, which computes hybrid embeddings that decompose captions into affirmative and negated components using a threshold $\tau{=}0.92$ derived from English data. We measure top-1 accuracy on the 4-way caption ranking task (chance = 25\%).

\section{Results}

\paragraph{Even multilingual VLMs exhibit significant negation gaps.} Table~\ref{tab:summary} summarizes baseline performance. CLIP shows the most extreme disparity: English achieves 39.3\% while Arabic (15.7\%), Tagalog (11.8\%), and Greek (18.0\%) fall \emph{below} chance, indicating systematic failure. SigLIP, trained on multilingual data, narrows this gap but still shows a 7.4-point spread. MultiCLIP achieves the most consistent performance (std.\ dev.\ 1.3\%), yet its mean accuracy of 41.2\% indicates that even the best multilingual model struggles with negation across the board.

\begin{table}[t]
\centering
\caption{\textbf{Baseline cross-lingual negation performance.} Even MultiCLIP, the most equitable model, achieves only 41\% mean accuracy, showing that multilingual training alone does not resolve negation understanding.}
\label{tab:summary}
\vspace{0.5em}
\small
\begin{tabular}{lccc}
\toprule
\textbf{Model} & \textbf{Mean Acc.} & \textbf{Std.\ Dev.} & \textbf{Max--Min Gap} \\
\midrule
CLIP & 23.5\% & 9.2\% & 27.5\% \\
SigLIP & 30.5\% & 2.6\% & 7.4\% \\
MultiCLIP & 41.2\% & 1.3\% & 4.4\% \\
\bottomrule
\end{tabular}
\end{table}

\paragraph{SpaceVLM as a diagnostic lens.} SpaceVLM produces substantial improvements for several languages, but the gains are not uniform and follow a typological pattern (Table~\ref{tab:spacevlm_delta}). Languages with negation expressed via independent particles \citep{miestamo2005}, such as English, Spanish, Greek, and Tagalog, show consistent large gains across all three models (+9.0 to +27.5pp). By contrast, languages with morphologically complex negation \citep{miestamo2005} (Russian, Arabic) or distinct negation particles with different semantic functions \citep{miestamo2005} (Chinese) show smaller or variable effects.

\begin{table}[t]
\centering
\caption{SpaceVLM reveals typological patterns ($\Delta$ = SpaceVLM $-$ Baseline). \textcolor{green!50!black}{Green}: gain $>$5pp; \textcolor{red}{Red}: decrease $>$5pp. Languages with adverbial negation benefit consistently; languages with complex morphological negation show variable effects, revealing that negation structure affects behavior.}
\label{tab:spacevlm_delta}
\vspace{0.5em}
\small
\begin{tabular}{l|ccc|ccc|ccc}
\toprule
& \multicolumn{3}{c|}{\textbf{CLIP}} & \multicolumn{3}{c|}{\textbf{SigLIP}} & \multicolumn{3}{c}{\textbf{MultiCLIP}} \\
\textbf{Language} & Base & Space & $\Delta$ & Base & Space & $\Delta$ & Base & Space & $\Delta$ \\
\midrule
English & 39.3 & 62.5 & \textcolor{green!50!black}{+23.3} & 34.6 & 54.6 & \textcolor{green!50!black}{+20.0} & 40.6 & 66.1 & \textcolor{green!50!black}{+25.5} \\
Chinese & 25.5 & 30.7 & +5.2 & 31.9 & 12.4 & \textcolor{red}{$-$19.5} & 42.5 & 38.6 & $-$3.9 \\
Arabic & 15.7 & 16.6 & +0.9 & 28.3 & 30.1 & +1.8 & 40.6 & 34.5 & \textcolor{red}{$-$6.1} \\
Greek & 18.0 & 33.9 & \textcolor{green!50!black}{+15.9} & 28.1 & 46.0 & \textcolor{green!50!black}{+17.9} & 38.9 & 66.4 & \textcolor{green!50!black}{+27.5} \\
Russian & 20.4 & 19.1 & $-$1.3 & 30.4 & 19.8 & \textcolor{red}{$-$10.6} & 43.3 & 36.3 & \textcolor{red}{$-$7.0} \\
Tagalog & 11.8 & 50.7 & \textcolor{green!50!black}{+38.9} & 27.2 & 31.0 & +3.8 & 41.6 & 50.6 & \textcolor{green!50!black}{+9.0} \\
Spanish & 33.6 & 56.2 & \textcolor{green!50!black}{+22.6} & 33.1 & 56.8 & \textcolor{green!50!black}{+23.7} & 41.3 & 67.5 & \textcolor{green!50!black}{+26.2} \\
\bottomrule
\end{tabular}
\end{table}

\paragraph{The pattern is robust.} We applied SpaceVLM to NegCLIP and ConCLIP—models fine-tuned specifically for negation on English data \citep{yuksekgonul2023when} and it produces the same typological pattern: strong English gains (57--69\%) with consistent gaps for Arabic (18--20\%), Russian (27--29\%), and Chinese (35--39\%). This persistence across five model-solution combinations suggests it reflects a genuine property of how these languages encode negation.

\section{Discussion: Implications for Equitable Deployment}

VLMs handle negation better in English than in other languages, which constitutes a clear disparity in model reliability across linguistic communities. 
This is especially concerning in safety-critical applications where differing results provide unequal service quality along linguistic lines. 
Our benchmark makes these disparities measurable. 
The typological pattern we observe also suggests that the \emph{way} a language expresses negation influences model behavior. 
Languages where negation is adverbial and structurally similar to English benefit from English-based corrections; languages with distinct negation systems do not. This implies that fairness audits for multilingual AI should consider the linguistic structures present in diverse languages and whether they adequately captured by models and their corrections. Addressing these gaps will likely require curating negation-rich multilingual pretraining data, developing tokenization strategies that preserve negation scope across morphological systems, and calibrating correction methods per typological group (e.g., tuning SpaceVLM's threshold $\tau$ per typological group). Additionally, because SpaceVLM's decomposition of captions into affirmative and negated components relies on parsing strategies suited to English syntax, structural differences across languages suggest that fine-tuning a multilingual LLM to more accurately extract affirmative and negated components could improve performance in a multilingual context.

\paragraph{Limitations.} Our translations relied on Google Translate with human verification of 30 samples per language (${\sim}$0.5\% of the dataset); fully human-verified translations would strengthen the benchmark. We evaluated SpaceVLM with its default English-optimized hyperparameters, and language-specific tuning of the threshold $\tau$ may reduce the cross-lingual variation we observe. Finally, we evaluated three open-source VLMs; extending to proprietary and domain-specific models (e.g., clinical VLMs for radiology) is an important direction for future work.

\section{Conclusion}

We present the first multilingual benchmark for negation understanding in VLMs, revealing that even explicitly multilingual models exhibit significant cross-lingual negation gaps. By applying SpaceVLM across seven typologically diverse languages, we uncover a pattern linking negation morphology to model behavior: languages with adverbial negation benefit consistently from corrections, while morphologically complex negation systems show variable responses. This difference shows that linguistic typology influences how models interpret meaning, and that those effects have real fairness implications. As VLMs are deployed globally, benchmarks that capture this typological diversity are essential for ensuring that alignment improvements benefit all linguistic communities, not just those whose languages happen to resemble English.

\bibliography{iclr2026_conference}
\bibliographystyle{iclr2026_conference}

\end{document}

%% file: math_commands.tex
\usepackage{amsmath,amsfonts,bm}

\def\eqref#1{equation~\ref{#1}}

\def\1{\bm{1}}

\DeclareMathAlphabet{\mathsfit}{\encodingdefault}{\sfdefault}{m}{sl}
\SetMathAlphabet{\mathsfit}{bold}{\encodingdefault}{\sfdefault}{bx}{n}



%% file: iclr2026_conference.bib
@article{alhamoud2025negation,
  title={Vision-language models do not understand negation},
  author={Alhamoud, Kumail and Alshammari, Shaden and Tian, Yonglong and Li, Guohao and Torr, Philip and Kim, Yoon and Ghassemi, Marzyeh},
  journal={arXiv preprint arXiv:2501.09425},
  year={2025}
}

@article{ranjbar2025spacevlm,
  title={{SpaceVLM}: Sub-space modeling of negation in vision-language models},
  author={Ranjbar, Sepehr Kazemi and Alhamoud, Kumail and Ghassemi, Marzyeh},
  journal={arXiv preprint arXiv:2511.12331},
  year={2025}
}

@inproceedings{radford2021clip,
  title={Learning transferable visual models from natural language supervision},
  author={Radford, Alec and Kim, Jong Wook and Hallacy, Chris and Ramesh, Aditya and Goh, Gabriel and Agarwal, Sandhini and Sastry, Girish and Askell, Amanda and Mishkin, Pamela and Clark, Jack and others},
  booktitle={ICML},
  year={2021}
}

@inproceedings{zhai2023siglip,
  title={Sigmoid loss for language image pre-training},
  author={Zhai, Xiaohua and Mustafa, Basil and Kolesnikov, Alexander and Beyer, Lucas},
  booktitle={ICCV},
  year={2023}
}

@inproceedings{carlsson2022mclip,
  title={Cross-lingual and multilingual {CLIP}},
  author={Carlsson, Fredrik and Eisen, Philipp and Rekathati, Faton and Sahlgren, Magnus},
  booktitle={LREC},
  year={2022}
}

@inproceedings{lin2014coco,
  title={Microsoft {COCO}: Common objects in context},
  author={Lin, Tsung-Yi and Maire, Michael and Belongie, Serge and Hays, James and Perona, Pietro and Ramanan, Deva and Doll{\'a}r, Piotr and Zitnick, C Lawrence},
  booktitle={ECCV},
  year={2014}
}

@inproceedings{blasi2022systematic,
  title={Systematic inequalities in language technology performance across the world's languages},
  author={Blasi, Damian and Anastasopoulos, Antonios and Neubig, Graham},
  booktitle={ACL},
  year={2022}
}

@inproceedings{yuksekgonul2023when,
  title={When and why vision-language models behave like bags-of-words, and what to do about it?},
  author={Yuksekgonul, Mert and Bianchi, Federico and Kalluri, Pratyusha and Jurafsky, Dan and Zou, James},
  booktitle={ICLR},
  year={2023}
}

@book{miestamo2005,
  title={Standard Negation: The Negation of Declarative Verbal Main Clauses in a Typological Perspective},
  author={Miestamo, Matti},
  year={2005},
  publisher={Mouton de Gruyter},
  address={Berlin}
}

@article{bella2023bridging,
  title={Towards Bridging the Digital Language Divide},
  author={Bella, G{\'a}bor and Helm, Paula and Koch, Gertraud and Giunchiglia, Fausto},
  journal={arXiv preprint arXiv:2307.13405},
  year={2023},
  doi={10.48550/arXiv.2307.13405}
}
